%% file: templateArxiv.tex
\documentclass{article}

\usepackage{PRIMEarxiv}
\usepackage[utf8]{inputenc} 
\usepackage[T1]{fontenc}    
\usepackage{url}            
\usepackage{booktabs}       
\usepackage{amsfonts}       
\usepackage{nicefrac}       
\usepackage{microtype}      
\usepackage{lipsum}
\usepackage{fancyhdr}       
\usepackage{graphicx}       
\graphicspath{{media/}}     

\usepackage{enumerate}
\usepackage{amsmath}
\usepackage{amssymb}
\usepackage{multirow}
\usepackage{xcolor}
\usepackage[pagebackref,breaklinks,colorlinks]{hyperref}
\usepackage[capitalize]{cleveref}
\crefname{section}{Sec.}{Secs.}
\Crefname{section}{Section}{Sections}
\Crefname{table}{Table}{Tables}
\crefname{table}{Tab.}{Tabs.}

\begin{document}

\title{Tag-based annotation creates better avatars}

\author{%
  Minghao Liu\\ UC Santa Cruz \\ \texttt{miu40@ucsc.edu} \\
  \and Zeyu Cheng \\ San Jose State University\\ \texttt{@outlook.com} \\
  \and Shen Sang \\ ByteDance Inc. \\ \texttt{shen.sang@bytedance.com} \\
  \and Jing Liu \\ ByteDance Inc. \\ \texttt{jing.liu@bydance.com} \\
  \and James Davis \\ UC Santa Cruz\\  \texttt{davis@cs.ucsc.edu} \\
}

\maketitle

\begin{abstract}
Avatar creation from human images allows users to customize their digital figures in different styles. Existing rendering systems like Bitmoji, MetaHuman, and Google Cartoonset provide expressive rendering systems that serve as excellent design tools for users. However, twenty-plus parameters, some including hundreds of options, must be tuned to achieve ideal results. Thus it is challenging for users to create the perfect avatar. A machine learning model could be trained to predict avatars from images, however the annotators who label pairwise training data have the same difficulty as users, causing high label noise. In addition, each new rendering system or version update requires thousands of new training pairs. In this paper, we propose a Tag-based annotation method for avatar creation. Compared to direct annotation of labels, the proposed method: produces higher annotator agreements, causes machine learning to generates more consistent predictions, and only requires a marginal cost to add new rendering systems.
\end{abstract}

\input{1_intro}
\input{2_related_work}

\input{3_method}
\input{4_results.tex}

\input{5_conclusion}

\newpage
{\small
\bibliographystyle{ieee_fullname}
\bibliography{egbib}
}
\end{document}

%% file: 1_intro.tex
\section{Introduction}

Well-designed avatar creation tools like Bitmoji ~\cite{bitmoji}, Google Cartoonset ~\cite{Google_cartoon}, and Metahuman ~\cite{metahuman} provide expressive tools for users to create digital figures based on themselves. However, customizing the ideal avatar  involves laborious selection and adjustment of parameters. Such a process consumes a significant amount of time from an average user without necessarily  resulting in their ideal design. Training a learning-based algorithm for avatar auto-creation is needed.

 Supervised learning requires the collection of pairwise training data. Given human face images, annotators manually create the corresponding avatars by selecting the best assets. Unfortunately, there are inherent issues with this \textit{direct} annotation method. During the creation process, some parameters such as hairstyle include hundreds of options with only minor differences. It is almost impossible for the annotators to consistently select a single optimal choice, resulting in low agreement with other annotators. When collected in this way, the dataset has high label noise, and majority vote aggregation does little to help. 

\input{Fig_tex/Fig_teaser}

Instead, we propose a \textit{Tag-based} annotation method for avatar creation. A semantically meaningful set of tags is designed as an intermediate representation which applies to both photographs and avatar renderings. Annotators label both training photographs and the set of stylized assets using these tags. For example, annotating \textit{hair length}, \textit{hair curly level}, and \textit{hair direction} instead of simply \textit{find the best option out of two hundred hairstyles}. This design provides better instructions and encourages annotators to search for more detailed features when labeling. The proposed Tag-based annotation results in higher annotator agreements.

Given a photograph, Tag annotations do not directly provide an answer for which asset is a best match. We use a search algorithm to evaluate the similarity of image tags and the tags of each possible asset. The asset with maximum similarity is selected as the best match. In addition to providing a best match, this search provides a score for each asset, so additional plausible options are available. In contrast, direct annotation produces a single best label with no information about the quality of any other asset.

In order to understand how this change in annotation style affects the final system, we compared supervised learning models trained on Tag-based labels with models trained on Direct labels. Experimental results show that models trained with the Tag-based system produces better and more consistent predictions. Example hairstyles predicted by the tag-based system are shown in Fig.~\ref{fig:teaser}.

Finally, we demonstrate that the Tag-based system helps with generalizability by showing results on several different avatar rendering systems. When shifting to a new rendering system, direct annotation requires a new large training set of human-avatar pairs. In contrast, the tag-based labels of human images can be reused, and the only new labels required are for the relatively small set of new assets. In a typical system, assets number in the hundreds while training photographs number in the tens of thousands, so the advantage is significant. 

This paper contributes a new method for avatar creation using \textit{tag-based} annotation. The advantages of this method include:
\begin{itemize}
    \item Cleaner labels with higher annotator agreement
    \item Better predictions from the ML model
    \item Lower cost generalization to new rendering systems
\end{itemize}






%% file: Fig_tex/Fig_teaser.tex
\begin{figure}[t]
\centering

    \includegraphics[width=0.99\linewidth]{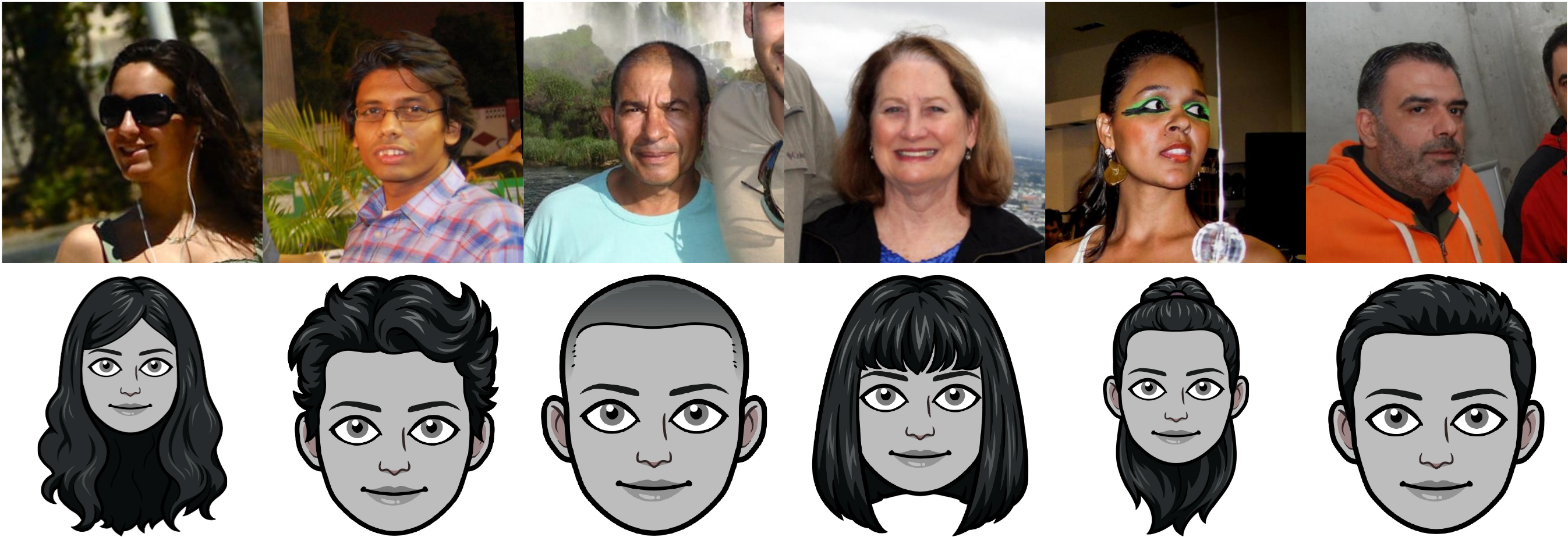}
    \caption{Avatar creation aims to automatically customize a stylized figure for a user. Hairstyle prediction is shown here because hair is the hardest attribute to pick automatically with hundreds of ambiguous options. We propose a Tag-based annotation method for avatar creation, which results in better labels (Sec \ref{sec: annotation better}), better predictions (Sec \ref{sec: model better}), and better generalizability (Sec \ref{sec: Generalizable}).}
    \label{fig:teaser}
    \vspace{-1.5em}
\end{figure}


%% file: 2_related_work.tex
\section{Related work}

\noindent \textbf{Image Stylization:}
Creating a virtual character from an input human portrait image needs to overcome the domain gaps between the real world and the target styles. Gatys et al.\cite{gatys2016image} matched feature information from CNN models to achieve style transfer. Cycle consistency loss was used for image-to-image transfer with paired-wise data supervision \cite{isola2017image} and with self-supervision \cite{zhu2017unpaired, park2020contrastive}. Recently, the development of GAN inversion methods \cite{richardson2021encoding, tov2021designing} results in excellent image decomposition and high-quality reconstructions, which has been applied to image stylization \cite{song2021agilegan, cao2018cari, zhu2021mind}. However, all of these methods focused on creating high-quality images in pixel space, as opposed to selecting assets in an avatar rendering system.

\vspace{0.4em}\noindent\textbf{Avatar creation using non-semantic parameters:}
Creating avatars in parameter spaces without semantic meaning has been well-studied for many years. Extremely high quality methods for photorealistic avatars using stereo vision ~\cite{beeler2010high, Yang_2020_CVPR}, and single input images  exist ~\cite{blanz1999morphable,peng2017parametric, deng2019accurate, xu2020deep, chen2021learning}, with multiple good survey papers ~\cite{egger20203d, zollhofer2018state}. 

Stylized avatar systems also exist. Some methods utilize sketches as the prior condition for generation ~\cite{han2017deepsketch2face,han2018caricatureshop}.  Other methods are guided by position and landmarks, extracting human facial features used to deform textures and meshes ~\cite{wu2018alive,cai2021landmark,lewiner2011interactive,vieira2013three}. Recently a conditional GAN has been applied in the generation process \cite{li2021deep,ye20213d}. However, these methods all utilized parameters without semantic meaning, making them inapplicable to avatar systems designed to provide user level customization of asset choice.

\noindent \textbf{Avatar creation using semantic parameters:}
To provide tools for customization of avatar creation, excellent rendering tools like Bitmoji~\cite{bitmoji}, Metahuman~\cite{metahuman}, and Google Cartoon Set~\cite{Google_cartoon} were created. These rendering systems provide explicit semantic meanings to each parameter and focus primarily on manual user creation. 

Avatar prediction has been explored using self-supervised methods to avoid the difficulty of manual labeling. When the avatar is semi-photorealistic, F2P \cite{F2P} utilizes neural imitators to mimic the behaviors of the rendering system, improving efficiency and applying textures for more photorealistic visual quality ~\cite{shi2020fast,lin2021meingame}. In the stylized domain, AgileAvatar ~\cite{AgileAvatar} introduced a domain transfer module to the avatar creation pipeline. However, these self-supervised methods rely heavily on carefully tuning each style. We provide a comparison to these methods in our results section. 

\noindent \textbf{Human face datasets:}
Training neural engines require the collection of human face datasets. FFHQ ~\cite{ffhq} provides a collection of high-quality human face images without annotation. CelebA ~\cite{CelebA} and MAAD\cite{maad1,maad2} datasets include some basic tags of facial attributes. FairFace ~\cite{FairFace} includes ethnic tags and provides a racially balanced set. 
Hairstyle specific datasets also exist. Figaro-1k ~\cite{svanera2016figaro} provides a limited set of samples, Hairstyle-30k ~\cite{yin2017learning} treats the task as an end-to-end classification task, while K-hairstyle ~\cite{kim2021k} focus on Korean hairstyles.
None of these datasets has labels matching the specific avatar rendering systems we use in our work. We make use of the FairFace dataset for photographs of human faces.



%% file: 3_method.tex
\section{Method}
Direct annotation for avatar creation and the challenges it introduces are discussed in Sec. \ref{sec: challenge}. Our tag-based annotation system is introduced in Sec. \ref{sec: data collection}. The search algorithm which relates tags to specific assets is discussed in 
Sec. \ref{sec:search algo}. The vision backbone and training loss we selected is provided in Sec.~\ref{sec: architectures}.

\input{Fig_tex/Fig_challenge}

\subsection{Direct annotation }
\label{sec: challenge}
Customizing a stylized avatar using rendering systems like Bitmoji requires users to tune more than twenty parameters. For example, hairstyle, face shape, and nose type. Some parameters include hundreds of options to choose from and we notice that some parameters dominate the perception of avatar quality. To enable detailed comparison between avatar annotation and prediction methods, we select a single parameter to demonstrate results in this paper. We choose hairstyle, because it is the attribute with the most variations, and assets often contain high ambiguity between options. Hairstyle also has high visual impact on an avatar, since it affects a large spatial region. We keep other parameters fixed to the extent possible in each rendering system we use.


Fig.~\ref{fig:challenge} shows some sample hairstyles from the Bitmoji rendering system with the grey-scaled, gender-neutral, default face. There are many hairstyles to choose from and direct annotation simply asks the annotator to pick the ground-truth asset class which best matches the input photograph. However decision making is complicated. Some options have very subtle differences between them. Often none are an ideal match and more than one hairstyle is a plausible option. A few examples of this ``no single best answer" phenomena are shown on the left side of the figure. Simply asking the annotators to directly label the best one out of two hundred options results in high label noise, and since annotators have low agreement majority vote aggregation does not often succeed.

\subsection{Tag-based annotation }
\label{sec: data collection}
In this paper, we propose Tag-based annotation. The goal is to map both human face images and avatar hairstyles to a tag space with semantic meaning. We defined our tags as in Tab.~\ref{tab:attrbutes}. Instead of providing the annotators with a massive number of options, we specifically ask them to annotate tag attributes from each region, for example \textit{Hair direction on the top of the head}, or \textit{Hair curliness level on the side of the head}. Detailed descriptions of each tag are included in the supplemental material.

\input{Fig_tex/tab_attrbutes}

Designing the appropriate tags to describe the hairstyle requires domain knowledge. Each tag requires a clear definition. For example,  \textit{short hair} and \textit{medium-short hair} is insufficient description for consistent labeling. We use an iterative design process to arrive at our final tag definitions. The researchers first designed tags to describe the hairstyles by simply looking at a set of human images and avatars. An annotator tagged all the avatar hairstyles using this tag design. A different annotator tagged a set of human images. Using the tags, the best matched avatar to each photo is retrieved. The researchers then evaluate the agreement between annotators, and the expressibility of the tag design, to make modifications to the set of tags. The process was repeated until tag design was considered sufficient.


After arriving at a tag designing, we perform the complete run of data annotation. Note that our tag design pipeline allows researchers to focus on iteratively improving their tag designs while not requiring them to work as annotators. By going through such a design process, researchers verify their designs, so that higher agreement between researchers and annotators is achieved.

\subsection{Search algorithm }
\label{sec:search algo}
Our designed tag system has 460,800 permutations, making it impossible to design a hairstyle for each permutation. This implies that for many human images there are no perfect hairstyle match.  To address this issue, we search through all existing hairstyles, computing tag similarity. The overall distance of a particular asset is computed as a weighted sum of individual tag distances. The weight of each attribute is listed in Tab.~\ref{tab:attrbutes}. To measure the tag distance for each attribute, we used L1 loss for continuous variables, and zero-one loss for discrete variables. 

The distance score from the search provides ranking information for all the hairstyles, while direct label annotation only provides the Top-1 result. Fig. \ref{fig:low_high_dist} shows visual samples of low and high-distance pairs. Note that the low-distance hairstyles have better visual similarity with the inputs, while the high-distance samples are visually dissimilar.

\input{Fig_tex/Fig_low_high_dist}

\subsection{ML training}
\label{sec: architectures}
We trained our learning-based models in a supervised manner. To extract feature information from the image, we used the open-source pre-trained Resnet-50~\cite{he2016deep} from the Pytorch~\cite{paszke2017automatic} library as our vision backbone and the initial training checkpoint. During training, we used L2 loss for continuous variables and cross-entropy loss for discrete classifications. 

%% file: Fig_tex/Fig_challenge.tex
\begin{figure*}[t]
\centering

  \includegraphics[width=0.99\linewidth]{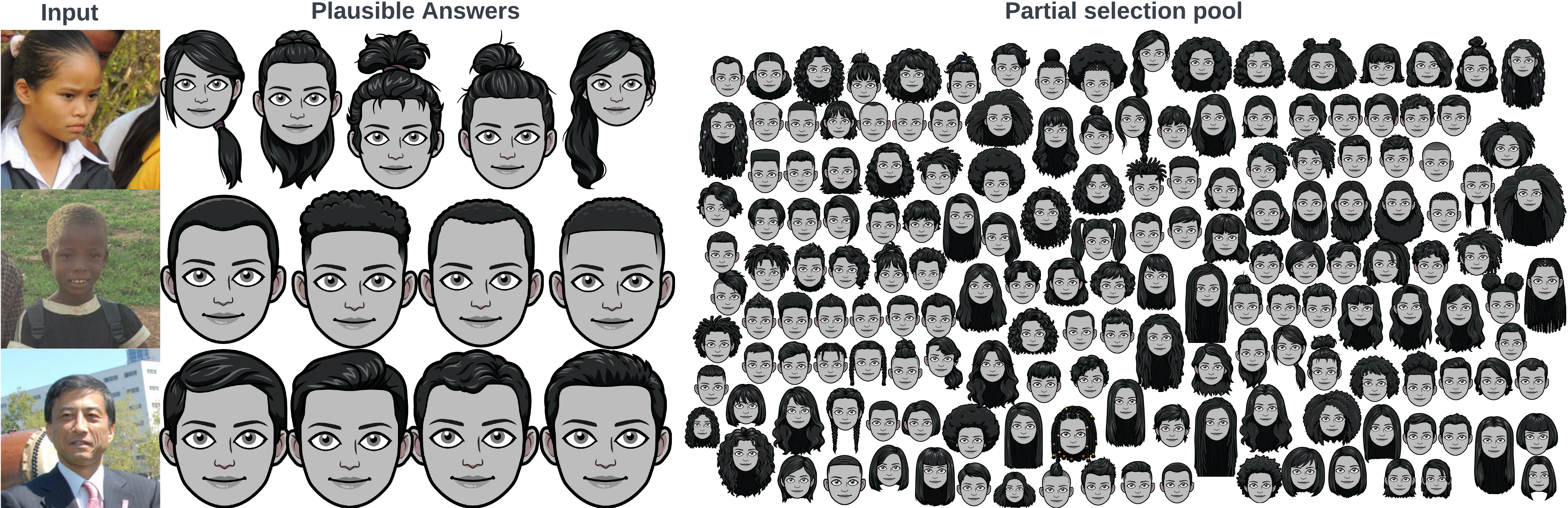}
  \caption{\textbf{Direct annotation is a challenging task: }There are hundreds of options in the selection pool, and for a given human portrait multiple plausible answers may exist. Each of the plausible answers may have only very minor differences, and none may be a perfect match for the image. This ambiguity leads to poor agreement between annotators.}
  
  \label{fig:challenge}
  \vspace{-0.7em}
\end{figure*}

%% file: Fig_tex/tab_attrbutes.tex
\begin{table}[t]
\centering

\scalebox{0.85}{
    \begin{tabular}{c|c|c|c|c}
    \hline
    \multirow{2}{*}{\textbf{Region}}        & \multicolumn{2}{c|}{\textbf{Annotation Tags}} & \multicolumn{2}{c}{\textbf{Distance calculation}}  \\  \cline{2-5}
                                   & Attributes     & \# Options    & Weight    & Type        \\ \hline
    \multirow{3}{*}{Top and front} & Length         & 6             & 2.25      & Continuous    \\ 
                                   & Direction      & 8             & 2         & Discrete         \\ 
                                   & Curly level    & 4             & 1         & Continuous       \\ \hline
    \multirow{2}{*}{On the side}   & Length         & 5             & 2.25      & Continuous    \\ 
                                   & Curly level    & 4             & 1         & Continuous       \\ \hline
    \multirow{4}{*}{Braid}         & Yes / No       & 2             & 5         & Discrete         \\ 
                                   & Count          & 4             & 2         & Discrete         \\ 
                                   & Position       & 3             & 1         & Discrete         \\ 
                                   & Type           & 5             & 1         & Discrete         \\ \hline
    \end{tabular}
}
\caption{\textbf{Attributes designed for Tag-based annotation.} The design process is described in Sec \ref{sec: data collection}. Additional detailed descriptions of each tag is included in the supplementary material.}
\vspace{-0.8em}
\label{tab:attrbutes}

\end{table}

%% file: Fig_tex/Fig_low_high_dist.tex
\begin{figure}[t]
\centering

  \includegraphics[width=0.99\linewidth]{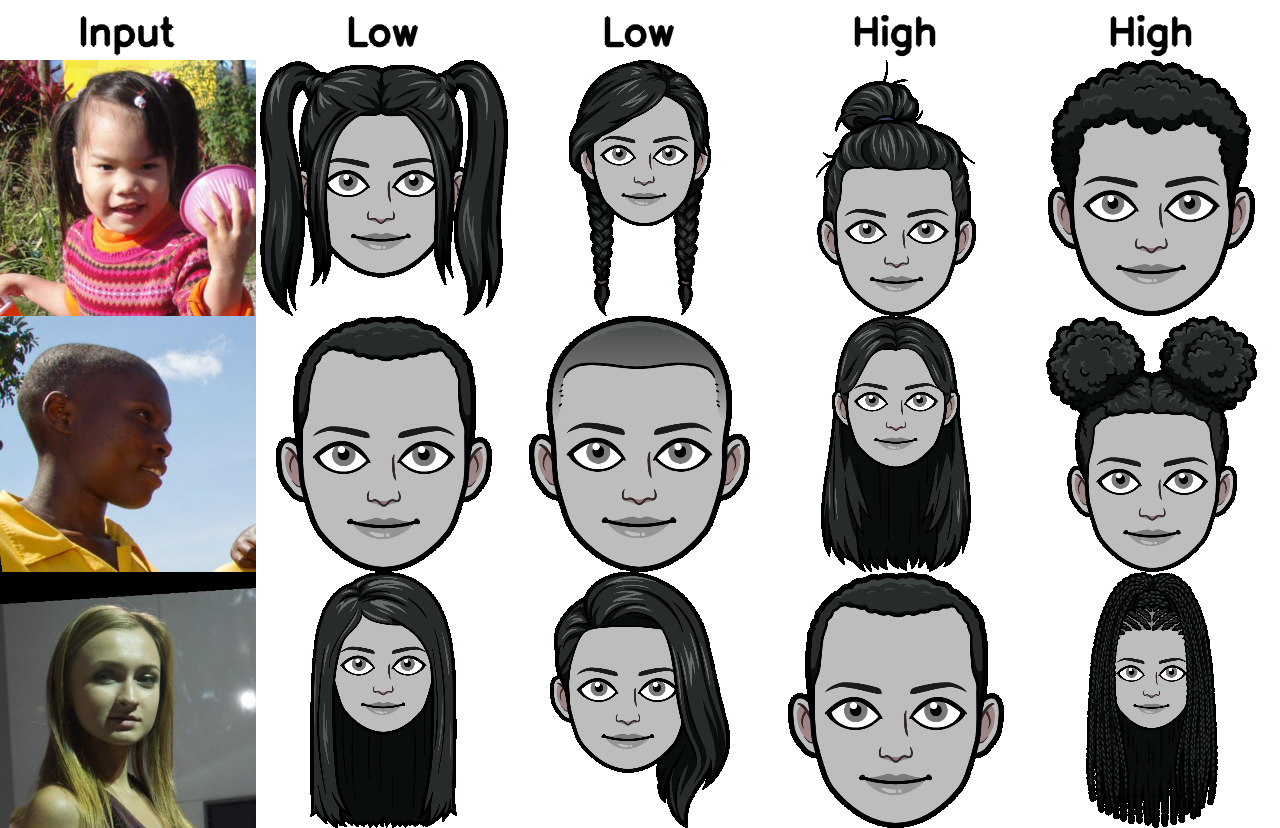}
  \caption{\textbf{Visualization of low and high distance samples:} The distance score is calculated by a search algorithm described in Sec~\ref{sec:search algo}, which measures the visual similarity between human images and avatar hairstyles. Low distance cases have good visual similarity to the person, while high distance cases appear visually dissimilar.}
  \vspace{-1em}
  \label{fig:low_high_dist}
\end{figure}

%% file: 4_results.tex
\section{Results and Experiments}
In this section, we demonstrate the advantage of Tag-based annotation with experimental results. Sec.~\ref{sec: annotation better} shows the advantage at the annotation level, Sec.~\ref{sec: model better} shows the advantage on model convergence and consistent predictions, and Sec.~\ref{sec: Generalizable} shows that a Tag-based system can easily be compatible with new rendering systems.

\subsection{Annotation Quality }
\label{sec: annotation better}
\input{Fig_tex/tab_annotation_compar}

\input{Fig_tex/Fig_annotation_compare}

\vspace{0.5em}\noindent
\textbf{Annotator Agreement:}
Label noise is a common problem in supervised learning models~\cite{cifar-n}. Collecting multiple copies of annotation for aggregation is often required to create a high-quality dataset. Using a majority vote is the most common way to aggregate labels and reduce label noise. However, given a large number of hairstyle options, there might exist multiple plausible answers, or alternatively, no ideal match and only partially correct answers. These situations cause low agreements between annotators.   Tab.~\ref{tab:annotation_compare} provides evidence of the severity of the problem when 3 annotators provide independent labels for each target. Agreement exists between annotators in only 31.0\% of cases when using direct annotation. In the majority of cases all three annotators provide different answers. On the other hand, annotator agreement on tags exists in 96.7\% of cases when using tag annotations. In order to fairly compare against direct annotation, we compute agreement on the Top-1 hairstyle chosen by each set of tag labels. Because multiple similar hairstyles exist, agreement is lower than on raw tag prediction at only 52.0\%, but this is substantially better (+21.0\%) than the agreement when labelers provide direct annotations. Unlike direct annotation, Tag-based annotation provides additional plausible answers which can be used to improve agreement and enable aggregation. Aggregating over the Top-4 matches from each of 3 labelers, reaches an agreement level of 80.39\% (+49.37\%).  We conclude that Tag-based labels are substantially more valuable in creating a ``clean" set of labels.


\vspace{0.5em}\noindent
\textbf{Annotation cost:}
The total cost of annotation is a function of labeling time for a individual image, and the total number of labels required. The cost of individual annotation is measured in time as shown in Tab.~\ref{tab:annotation_compare}. We asked annotators from two different skill levels to annotate images and recorded annotation time. Our experiment shows well-trained professional annotators need marginally less time to annotate a face when using the Tag-based system. In contrast, untrained workers obtained from Amazon Mechanical Turk need more time when using tag-based labels. In either case, the differences in individual labeling time are bounded and not the most significant factor.

Direct annotation requires a completely new set of labels each time a change is made to the set of available hairstyles, with associated retraining of the prediction model given the new labels. In contrast, tag-based systems only need to label the new hairstyles, with no new tag labels on the much larger set of training images, and no retraining of the tag predictor. Since most avatar rendering systems will be updated with new artistic assets occasionally, we conclude that there is a substantial savings on label cost using tag-based annotation.

\vspace{0.5em}\noindent
\textbf{Visual Quality comparison:}
To compare the visual quality between direct and Tag-based annotation, we conduct two user studies through Amazon Mechanical Turk~\cite{mturk}: \textit{Matching} and \textit{Preference}. In the \textit{Matching} task, we evaluate whether visual similarity is sufficiently close that evaluators can tell which avatar goes with which human. A human image is shown with the corresponding avatar hairstyle and three random distractor hairstyles. The evaluator is required to match the human image and avatar image and is scored as a correct match if the evaluator correctly picks the original annotation. A high matching score indicates the avatar represents the human well. A total of 1,224 judgments were collected. As the result shows in Tab.~\ref{tab:annotation_compare}, our proposed Tag-based annotation preserves user identity marginally better compared to direct annotation. In the \textit{Preference} task, avatar results from both methods are presented for comparison with the human image. The evaluators were asked to provide their preferences by choosing one of the results or indifferent, a total of 612 judgments were collected. The results showed a precise split of 306 judgements for each method. The combined results of both studies indicate that our tags are sufficiently expressive to act as a replacement for direct annotation in terms of visual quality. 

A visual comparisons of the two annotations methods is provided in Fig.~\ref{fig:annotation_compare}. Both annotation methods result in plausible answers. In addition to the Top-1 match, a Tag-based system can provide ranking information for all other hairstyles. The figure also shows Top-5 results from the Tag-based annotations. The visual quality of these results is reduced, but they remain plausible.

\subsection{ML Prediction Quality}
\label{sec: model better}
Tag-based labels result in better ML models when used for selecting avatar assets, in terms of both visual quality and consistency. 

\subsubsection{Baseline methods and dataset}
\label{sec: baseline}
We compare several methods to understand the effects of tag-based annotation.
 We choose two state of the art baseline methods using self-supervised models, a model supervised by direct labels, and a model supervised by Tags. 

\vspace{0.5em}\noindent
\textbf{Self-supervised baselines}
\begin{itemize}
    \item \textit{F2P}~\cite{F2P}: an optimization-based method designed for realistic game character creation. 
    \item \textit{AgileAvatar}~\cite{AgileAvatar}: a state-of-the-art learning-based method for stylized avatar creation.  
\end{itemize}

\vspace{0.5em}\noindent
\textbf{Supervised baselines:}
Both supervised baselines use an identical vision back bone with similar loss functions. 
\begin{itemize}
    \item \textit{Direct pred}: Treat the task as a classification task, predict the best hairstyle, and calculate training loss against direct annotation. 
    \item \textit{Tag pred}: Treat the task as a classification task, predict the tags, and calculate training loss against the Tag-based annotations. The final hairstyle prediction is determined using the search described in Sec.~\ref{sec:search algo}. 
\end{itemize}

\vspace{0.5em}\noindent
\textbf{Dataset and annotation:}
We used human face images from the FairFace ~\cite{FairFace} dataset. Compared to the commonly used CelebA ~\cite{CelebA} dataset, FairFace is racially balanced, which ensures our trained model works well on diverse faces. FairFace also includes blurry images, which requires the trained model to be robust on lower-quality images.

\vspace{0.5em}\noindent
\textbf{Annotation:}
We collected Tag-based annotation for 17k images from FairFace using professional annotators. We randomly choose 14.5k for training, 2.8k for testing, and 204 images as a hold out set for human evaluation studies.

Inter-annotator agreement is higher for tag-based annotation than direct annotation. To avoid bias due to the quality of annotations collected, we create a set of direct labels using the tag-based labels. We use the tags for each image to find the single best hairstyle. We treat these hairstyles as the training targets for \textit{Direct Pred}. Thus both supervised learning methods share the same Top-1 training targets.

\subsubsection{Better prediction quality}
\input{Fig_tex/Fig_model_compare}
\input{Fig_tex/tab_model_matching_scores.tex}


\textbf{Visual comparisons for models:}
We visually compare all four learning-based methods in Fig.~\ref{fig:model_compare}. F2P~\cite{F2P} was designed for realistic human-style avatars and fails frequently on stylized avatars. AgileAvatar~\cite{AgileAvatar} has a human stylization module to overcome the style domain gap and has significant improvements over F2P;  however it is still not as good as the supervised learning methods. Comparing the \textit{Direct pred} and \textit{Tag pred} models, training a model using the Tag-based annotation helps the model to capture the detailed features of the hairstyles. For example, capturing the double pony-tail for the girl in the third row, and the curliness of the hair for the man in the sixth row. 

\vspace{0.5em}\noindent
\textbf{Numerical comparison for models:}
 We conducted user studies to support the visual comparison with results shown in Tab.~\ref{tab:model matching scores}. The Matching test requires evaluators to match a randomly selected human against 4 avatar results. Such a test measures how well the model captures the human identity. If the avatar looks very similar to the photo, then identifying it is easy. If the avatar does not look similar it is hard to distinguish from randomly chosen distractors. Higher matching scores are better. F2P is designed for photorealistic avatars and performs poorly. The carefully tuned self-supervised AgileAvatar reaches a similar matching score to supervised \textit{Direct pred}, only 3.57\% lower. However, Tag-based annotation performs best, preserving more user identity than all other baselines.

\subsubsection{Better convergence, More consistent predictions }
\label{sec:converge consistent }
\input{Fig_tex/Fig_consistent_compare}
\input{Fig_tex/tab_consistent_compare}

\textbf{Convergence:}
During training, both supervised learning models have the same best-matching hairstyle for each human image. In the case of direct prediction this hairstyle is explicitly used in the target loss, while in the case of tag prediction it is implicitly defined by the target tags. Higher accuracy on the test set indicates converging to a more optimal point. Tab.~\ref{tab:model consistent} provides prediction accuracies. Naive comparison shows that direct prediction finds correct answer 10.29\% of the time while tag prediction is correct 95.72\% of the time. However this is not an entirely fair comparison because direct prediction must choose among hundreds of classes, while tag prediction has many fewer classes available. To create a fair comparison we require both methods to choose a final hairstyle among the hundreds available and compare this to the ``ground-truth" target. We compare model convergence toward the target loss using Top-K accuracy.  Training the model on the Tag-based systems, results in a better Top-1 {\small{\tt{\textcolor{red}{(+6.87\%)}}}} and Top-5 {\small{\tt{\textcolor{red}{(+9.32\%)}}}} accuracy compared to model training on direct labeling.  

The low Top-1~(17.16\%) accuracy seems to contradict our claim of good quality predictions in Tab~\ref{tab:model matching scores}; however, in reality, Top-1 accuracy only tells a partial story. There are hundreds of hairstyles, and some options have very minor difference from other options. Given the potential existence of multiple correct answers, it's not always wrong to predict something other than the ``ground truth" option. Thus Top-1 and Top-5 are good indicators of only the relative quality of two prediction models, not the absolute quality.

As an alternative metric for quality, we used our search algorithm described in Sec.~\ref{sec:search algo} to calculate the distance between the human face and the final avatar predictions, using the tags originally provided by annotators. A low distance score indicates greater similarity between the human face and avatar at the feature level. The result in Tab.~\ref{tab:model consistent} shows \textit{Tag pred} has a lower average distance for the Top\nobreakdash-1(2.51) and Top-5(3.16) predictions. Note that most human face images don't have a perfectly matching avatar so a distance score of 0.0 is impossible. We include the distance for ``ground truth" hairstyles as chosen by annotators as an indication of a lower bound on prediction quality.  

\vspace{0.5em}\noindent
\textbf{Consistency:}
We visual the Top-5 predictions of both supervised learning methods in Fig.~\ref{fig:consistent_compare}. While both methods produce plausible Top-1 answers, the Top-5 predictions from \textit{Tag Pred} have better consistency compared to \textit{Direct Pred}. The \textit{Direct Pred} model treats each hairstyle as an independent class without considering their visual similarities to the human image. \textit{Tag pred}, on the other hand, is trained to predict the visual features defined by human researchers. Thus the model was encouraged to focus on the features that are important to human observers, thus resulting in consistent Top-K predictions. Notice for example that even when direct prediction correctly predicts a short hairstyle as the Top-1 result, the next best prediction might be long hair.

\subsection{Generalizability}
\label{sec: Generalizable}

\input{Fig_tex/Fig_other_sys}
Annotating a dataset costs time and effort from both researchers and annotators. In our case, about 17k sets of tags for human images and avatar hairstyles. To make our trained model compatible with any other rendering system, the model requires an entirely new set of 17k labels if using direct annotation. 

Using a Tag-based system, on the other hand, can significantly reduce this cost. Since the tags for the human images are independent of rendering systems, human tags can be reused. We are only required to collect new tags for the avatar assets in the new rendering systems. Most rendering systems provide ``only" 200 hairstyle variations, costing less than 2\% of relabeling the human images. In addition the original tag prediction ML model remains valid, and does not need to be retrained.

To demonstrate the generalizability of the Tag-based system, we collected tags for avatar samples from Bitmoji~\cite{bitmoji}, Google Cartoonset \cite{Google_cartoon}, MetaHuman ~\cite{metahuman}, and NovelAI ~\cite{novelai}. For all these systems, we only controlled the hairstyles:
\begin{itemize}
    \item \textbf{Bitmoji:} A cartoon avatar rendering system. We used the gender-neutral, gray-scaled default faces.
    \vspace{-0.5em}
    \item \textbf{Google Cartoon set:} A dataset of the results from a cartoon avatar rendering system. We controlled the hairstyles and left other attributes set random. 
    \vspace{-0.5em}
    \item \textbf{Metahuman:} A realistic human avatar rendering system. Hairstyle is varied while leave other parameters constant. We choose between males and female default avatars using the gender tags in the FairFace dataset. 
    \vspace{-0.5em}
    \item \textbf{NovelAI:} A diffusion-based text-to-image generation model specialized for cartoon-style figures. We asked artists to select text prompts to generate cartoon figures with different hairstyles and treat the resulting images as a set of avatars.
\end{itemize}

Fig.~\ref{fig:other_sys} shows the model-predicted visual results. Given the tag predictions from our supervised learning model, the search algorithm finds the closest matching hairstyle in each system based on the avatar tags. Not every system contains every hairstyle so some matches are closer than others, but overall the avatars selected are good approximations of the input photograph.

%% file: Fig_tex/tab_annotation_compar.tex
\begin{table}[t]
\centering

\scalebox{0.73}{

\begin{tabular}{c|ccc}
\toprule
                            &                             & Direct annotation       & Tag-based annotation \\ \hline

\multirow{5}{*}{\begin{tabular}[c]{@{}c@{}}Chance \\ Agreement \\ exists\end{tabular}}  & Tag level                   & NA                      & \textbf{96.7\%}        \\ \cline{3-4} 
                            & Final Top-1                  & 31.0\%                    & \textbf{52.0\%}        \\ 
                            & Final Top-2                  & NA                      & \textbf{67.2\%}        \\  
                            & Final Top-3                  & NA                      & \textbf{73.5\%}        \\ 
                            & Final Top-4                  & NA                      & \textbf{80.3\%}        \\ \hline
\multirow{2}{*}{Time}       & Skilled annotators          & 25.1s                  & \textbf{23.6s}        \\ \
                            & Random Turker               & \textbf{48.4s}            & 112.6s         \\ \hline
\multirow{2}{*}{User study} & Matching                    & 89.4\%                 & \textbf{92.7\%}        \\ 
                            & Preference                  & \multicolumn{2}{c}{306 :  306}            \\ 
\bottomrule
\end{tabular}

}
\caption{\textbf{Numerical comparisons of annotation methods: } Agreement between annotators is a measure of quality. It is also correlated with the chance that majority vote aggregation helps reduce label noise. Tags have much higher agreement than direct labels. Using tags to predict the Top-1 hairstyle also produces better agreement than asking annotators to directly label the best hairstyle. Tag-based annotation additionally provides multiple Top-K matches for each annotator, where direct annotation only results in the Top-1 option. The agreement level increased to 80.3\% {\small{\tt\textcolor{red}{(+49.3\%)}}} by aggregating answers from the Top-4 matches from Tag-based annotation.  Time analysis shows a skilled annotator spent a similar amount of time to annotate a face using both annotation methods. A user study showed that Tag-based annotation results in similar visual quality. }
\label{tab:annotation_compare}
\vspace{-1.3em}
\end{table}

%% file: Fig_tex/Fig_annotation_compare.tex
\begin{figure}[t]
\centering

  \includegraphics[width=0.80\linewidth]{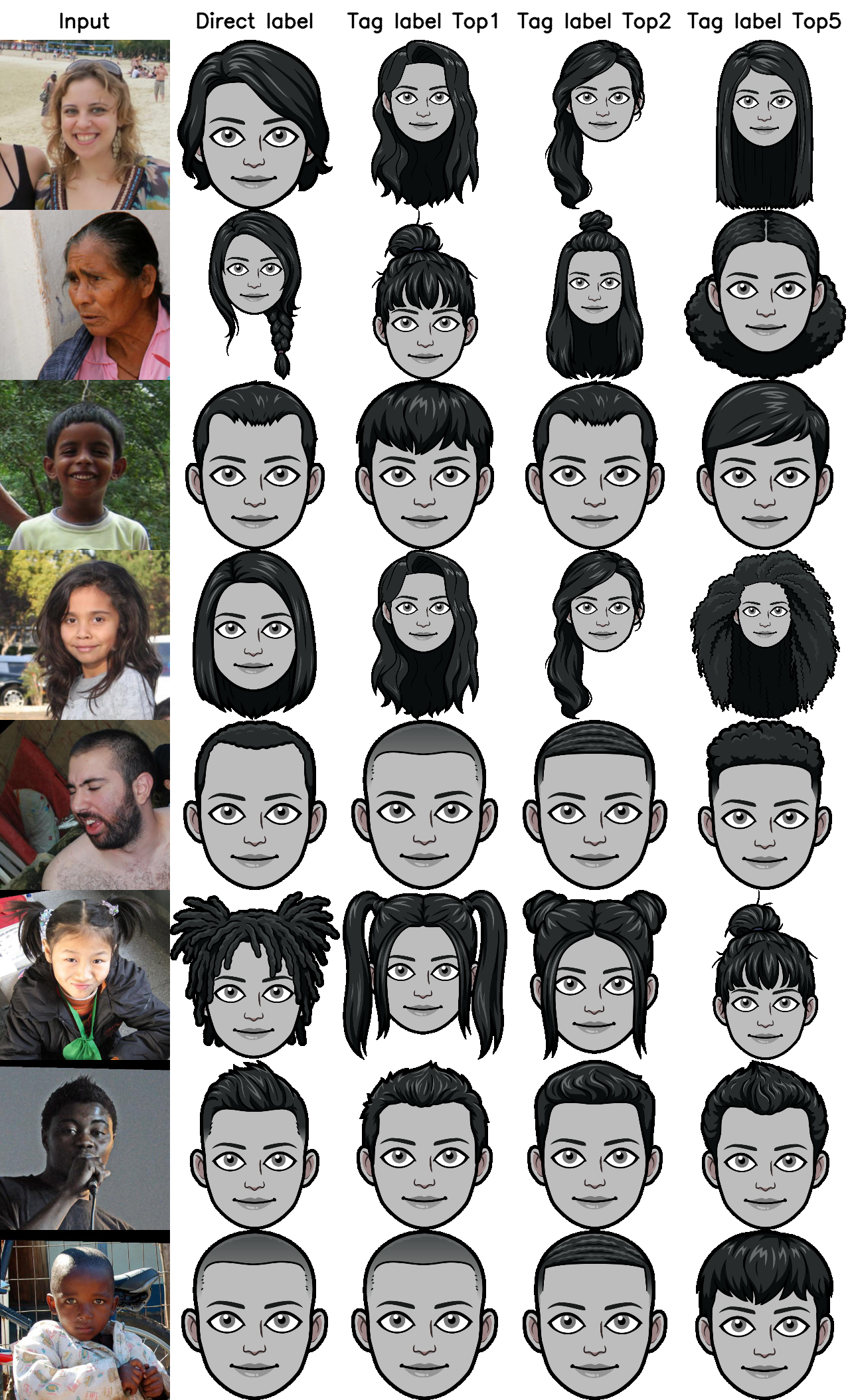}
  \caption{\textbf{Visual comparison of annotation results:}  Both direct and Tag-based annotation methods provide plausible answers, however Tag-based annotation provides additional backup recommendations.}
  \label{fig:annotation_compare}
  \vspace{-1.3em}
\end{figure}

%% file: Fig_tex/Fig_model_compare.tex
\begin{figure}[t]
\centering

  \includegraphics[width=0.75\linewidth]{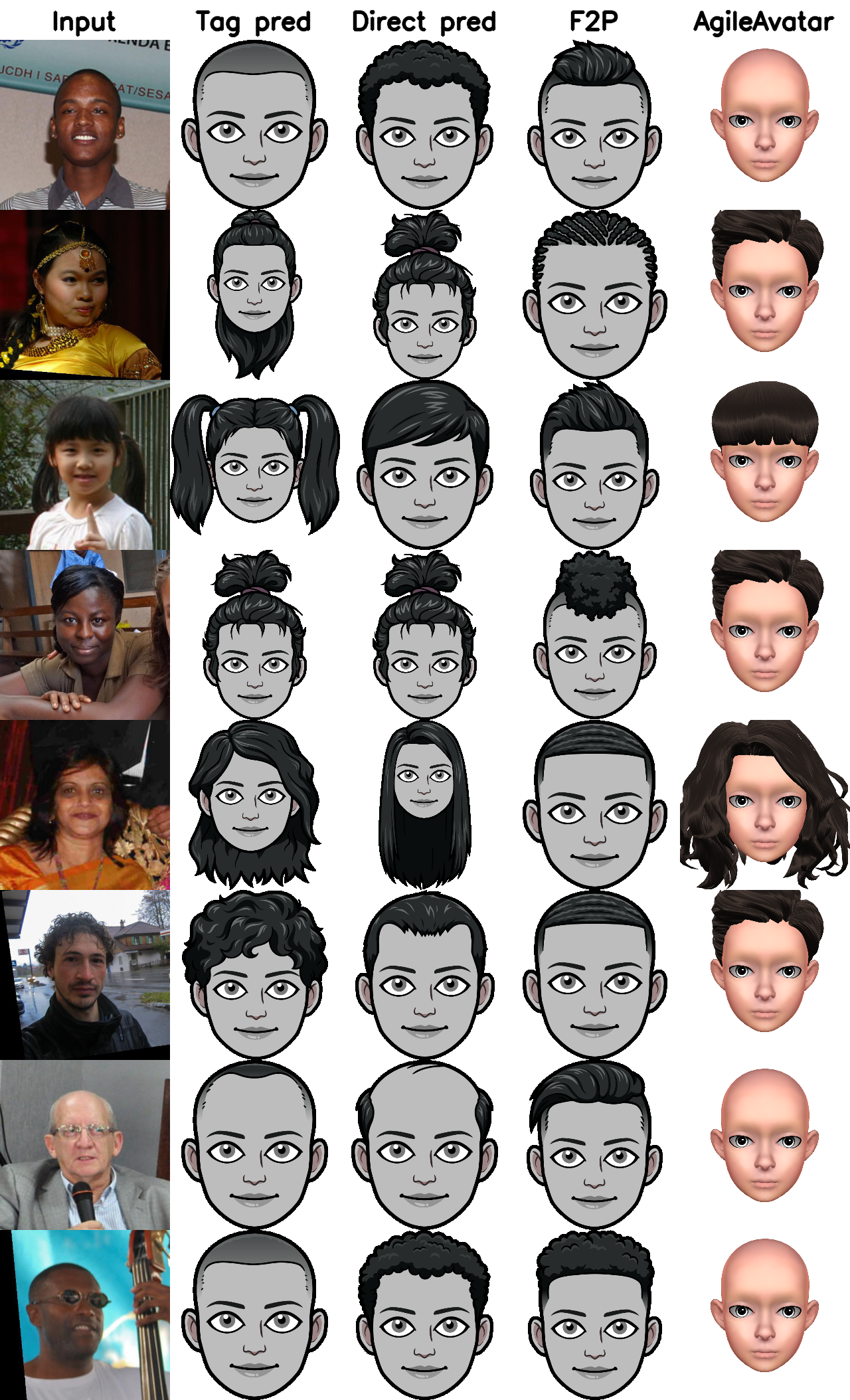}
  \caption{\textbf{Visual comparisons of model predictions: } \textit{Tag pred} and \textit{Direct pred} are supervised learning methods, which results in better visual avatars compared to self-supervised learning models F2P\cite{F2P} and AgileAvatar\cite{AgileAvatar}. F2P was designed for photorealistic avatar creation and fails to produce good results on stylized avatars. AgileAvatar produces better results with the help of stylization modules, though it captures less feature information than supervised learning methods. }
  \label{fig:model_compare}
  \vspace{-0.3em}
\end{figure}

%% file: Fig_tex/tab_model_matching_scores.tex
\begin{table}[t]
\centering

\scalebox{0.99}{
\begin{tabular}{l|ll}
\toprule
                                 & Trained methods   & Matching \\ \hline
\multirow{2}{*}{Self-supervised} & F2P\cite{F2P}     & 34.73\%  \\ 
                                 & AgileAvatar \cite{AgileAvatar} & 67.22\%  \\ \hline
\multirow{2}{*}{Supervised (ours)}      & Direct pred       & 70.79\%  \\ 
                                 & Tag pred          & \textbf{83.92\%}  \\ 

\bottomrule
\end{tabular}

}

\caption{\textbf{Quality comparisons of learning methods:} A matching test asks evaluators to identify the best avatar when the ML produced avatar is presented with 3 alternates as distractors.  A higher matching score indicates better representations of the input human images. Tag-based prediction performs better than existing self-supervised state of the art, and in comparison to direct annotations.}
\label{tab:model matching scores}
\vspace{-1em}
\end{table}

%% file: Fig_tex/Fig_consistent_compare.tex
\begin{figure}[t]
\centering
    \includegraphics[width=0.75\linewidth]{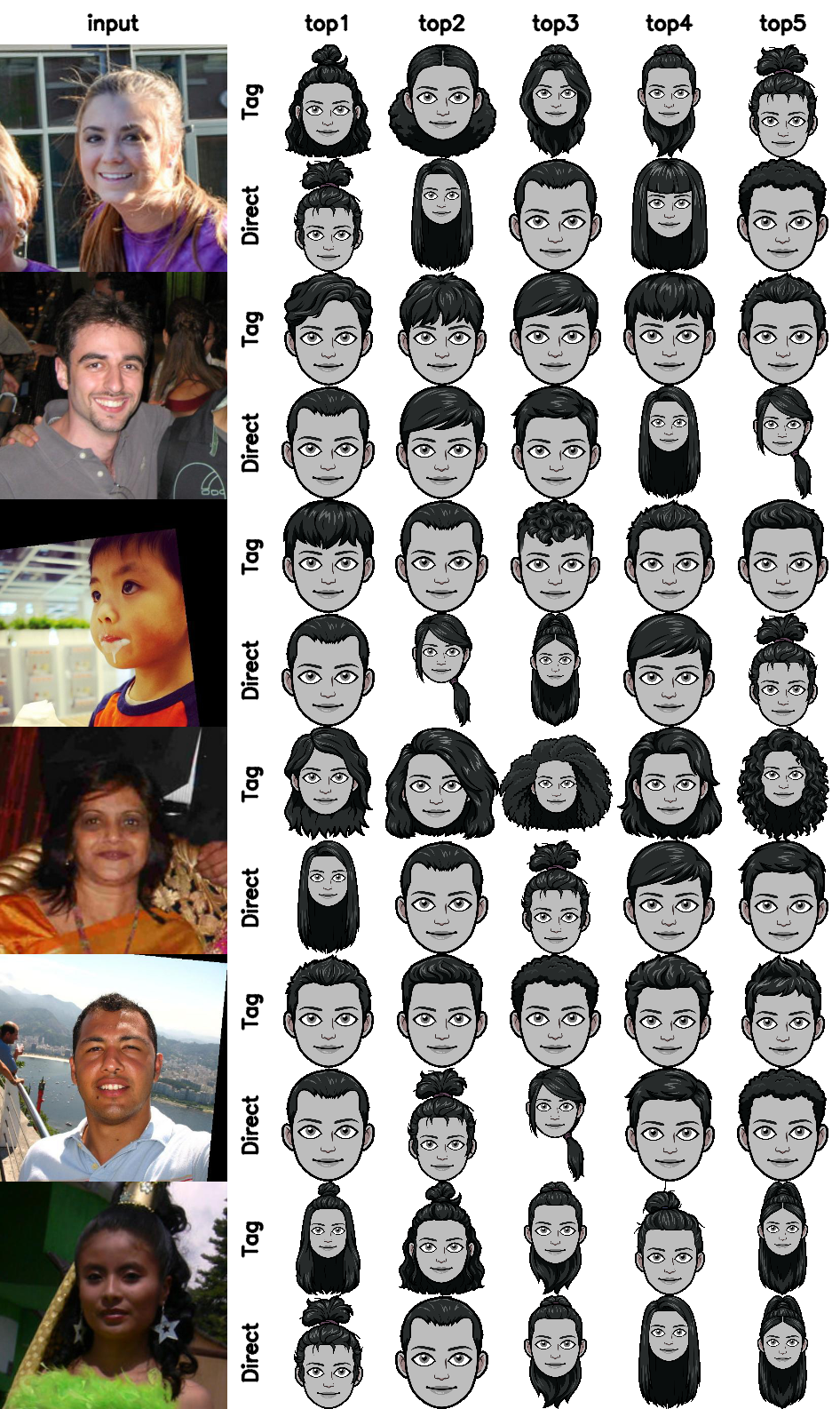}
    \caption{\textbf{Prediction consistency: }Visual comparisons of the Top-5 predictions from supervised learning models trained using tag-based and direct annotations. While both methods produce plausible Top-1 answer, The Top-5 predictions from \textit{Tag Pred} have better consistency than those from \textit{Direct Pred}. Notice for example that direct prediction sometimes includes both short and long hair in its top five predictions.}
    
    \label{fig:consistent_compare}
    \vspace{-1em}
\end{figure}

%% file: Fig_tex/tab_consistent_compare.tex
\begin{table}[t]
\centering

\scalebox{0.75}{
\begin{tabular}{l|cc|c|cc}
\toprule
            & \begin{tabular}[c]{@{}c@{}}Top-1 \\Accuracy\end{tabular}  & \begin{tabular}[c]{@{}c@{}}Top-5 \\Accuracy\end{tabular} & \begin{tabular}[c]{@{}c@{}}Tag pred \\ Accuracy\end{tabular} & \begin{tabular}[c]{@{}c@{}}Distance \\ Top-1\end{tabular} & \begin{tabular}[c]{@{}c@{}}Distance \\ Top-5\end{tabular} \\ \hline
Direct pred & 10.29  \% & 32.84 \% & NA                                                      & 6.09          & 8.25                                                        \\ \hline
Tag pred    & \begin{tabular}[c]{@{}c@{}}\textbf{17.16\% }\\{\tt\small{\textcolor{red}{(+6.87\%)}}}\end{tabular} & \begin{tabular}[c]{@{}c@{}}\textbf{42.16\% }\\ {\tt\small{\textcolor{red}{(+9.32\%)}}}\end{tabular} & \textbf{95.72\%}                         & \textbf{2.51}          & \textbf{3.76}                                                       \\ \hline
Annotation      &           &          &                                                         & 2.29          & 3.39                                                        \\ 
\bottomrule
\end{tabular}
}
\caption{\textbf{Convergence comparisons: } Tag annotations can be used to build better ML models. Using the final avatar target to calculate Top-1 and Top-5 accuracy for each supervised learning method shows a better convergence on the \textit{Tag-pred} model. The distance score is an alternate measure of similarity between a human face and avatar, and also shows better performance for tag-based prediction. }
\label{tab:model consistent}
\vspace{-2.3em}
\end{table}

%% file: Fig_tex/Fig_other_sys.tex
\begin{figure}[t]
\centering

  \includegraphics[width=0.75\linewidth]{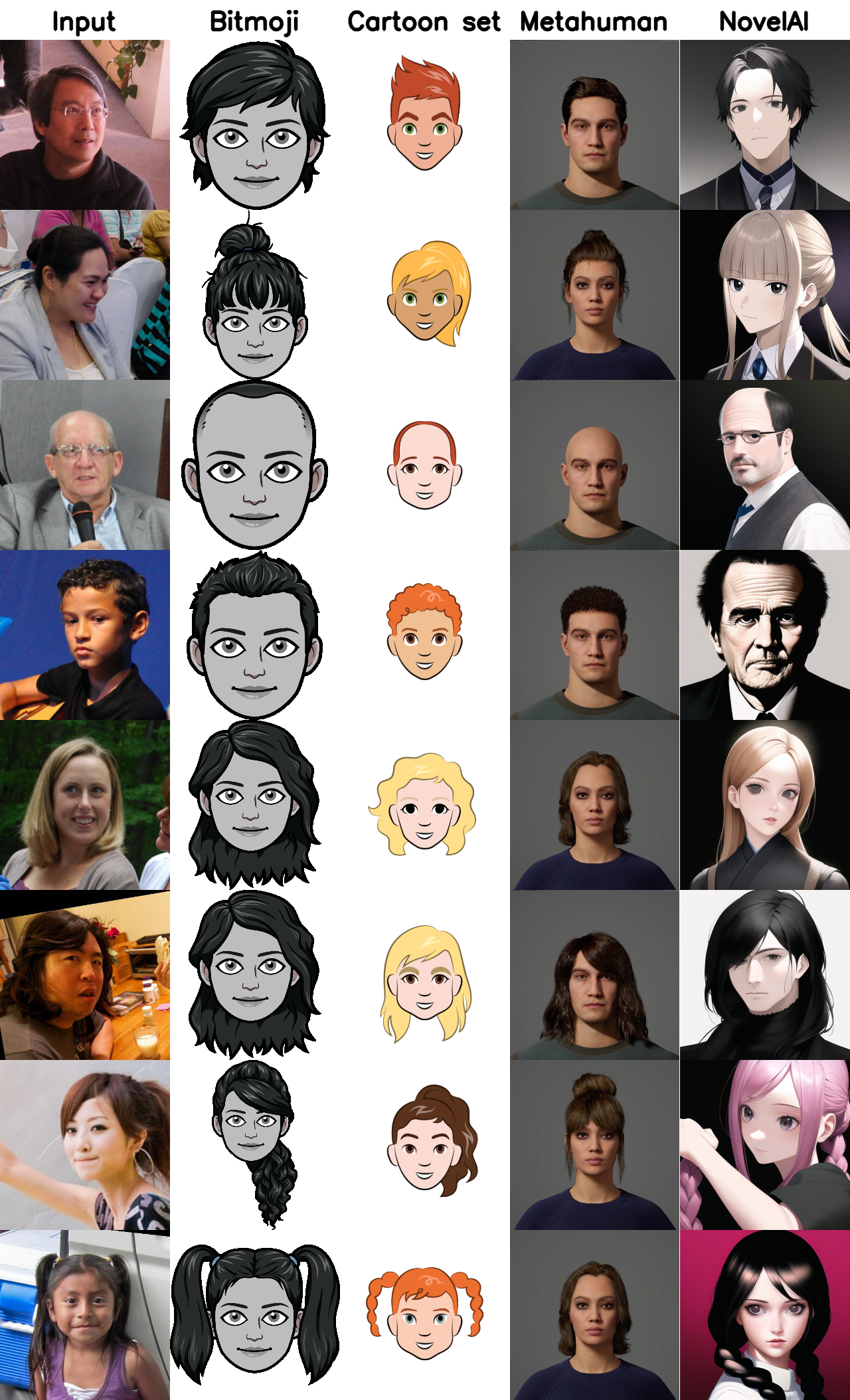}
  \caption{\textbf{Easily expandable to other systems: }Models trained on Tag-based annotation can easily be reused with other rendering systems. The tags originally predicted for use with the Bitmoji system were used to select avatars in three alternate systems.}
  \label{fig:other_sys}
\end{figure}

%% file: 5_conclusion.tex
\section{Limitations and conclusion}


\noindent\textbf{Limitations:}
To benefit from the Tag-based system, a set of carefully designed tags with clear definitions is required, along with the cost in time and effort to produce it.

The search algorithm is sensitive to tag prediction errors. While tag prediction is substantially more accurate than direct prediction, a wrong prediction will mislead the search. 

As with all avatar prediction methods, our work lacks a universally accepted benchmark or evaluation metric. Commonly used L2 loss, perceptual loss\cite{johnson2016perceptual}, or Top-K accuracy do not well represent actual user preference. This paper conducted user studies for result evaluations, but this also has shortcomings since evaluators are not the same as actual users. 

\vspace{0.5em}
\noindent\textbf{Conclusion:}
In summary, we present a Tag-based annotation method for avatar creation. Our approach results in higher annotation quality with higher annotator agreements. Supervised learning models guided by the proposed annotation converge to a more optimal point, and output more consistent predictions. Our proposed method can easily be applied to new rendering systems with marginal cost. Experimental results including numerical, visual, and user studies support our claims. 

%% file: templateArxiv.bbl
\begin{thebibliography}{10}\itemsep=-1pt

\bibitem{mturk}
Amazon mechanical turk.
\newblock \url{https://www.mturk.com}.
\newblock Accessed: 2022-11-10.

\bibitem{beeler2010high}
Thabo Beeler, Bernd Bickel, Paul Beardsley, Bob Sumner, and Markus Gross.
\newblock High-quality single-shot capture of facial geometry.
\newblock In {\em ACM SIGGRAPH 2010 papers}, pages 1--9. ACM New York, NY, USA,
  2010.

\bibitem{bitmoji}
Bitmoji.
\newblock \url{https://www.bitmoji.com/}.
\newblock Accessed: 2022-11-10.

\bibitem{blanz1999morphable}
Volker Blanz and Thomas Vetter.
\newblock A morphable model for the synthesis of 3d faces.
\newblock In {\em Proceedings of the 26th annual conference on Computer
  graphics and interactive techniques}, pages 187--194, 1999.

\bibitem{cai2021landmark}
Hongrui Cai, Yudong Guo, Zhuang Peng, and Juyong Zhang.
\newblock Landmark detection and 3d face reconstruction for caricature using a
  nonlinear parametric model.
\newblock {\em Graphical Models}, 115:101103, 2021.

\bibitem{cao2018cari}
Kaidi Cao, Jing Liao, and Lu Yuan.
\newblock Carigans: Unpaired photo-to-caricature translation, 2018.

\bibitem{chen2021learning}
Zhixiang Chen and Tae-Kyun Kim.
\newblock Learning feature aggregation for deep 3d morphable models.
\newblock In {\em Proceedings of the IEEE/CVF Conference on Computer Vision and
  Pattern Recognition}, pages 13164--13173, 2021.

\bibitem{Google_cartoon}
Forrester Cloe, Inbar Mosseri, Dilip Krishnan, Aaron Sarna, Aaron Maschinot,
  Bill Freeman, and Shiraz Fuman.
\newblock Cartoon set, 2022.
\newblock data retrieved from Google Machine Perception organization,
  \url{https://google.github.io/cartoonset/people.html}.

\bibitem{deng2019accurate}
Yu Deng, Jiaolong Yang, Sicheng Xu, Dong Chen, Yunde Jia, and Xin Tong.
\newblock Accurate 3d face reconstruction with weakly-supervised learning: From
  single image to image set.
\newblock In {\em Proceedings of the IEEE/CVF Conference on Computer Vision and
  Pattern Recognition Workshops}, pages 0--0, 2019.

\bibitem{egger20203d}
Bernhard Egger, William~AP Smith, Ayush Tewari, Stefanie Wuhrer, Michael
  Zollhoefer, Thabo Beeler, Florian Bernard, Timo Bolkart, Adam Kortylewski,
  Sami Romdhani, et~al.
\newblock 3d morphable face models—past, present, and future.
\newblock {\em ACM Transactions on Graphics (TOG)}, 39(5):1--38, 2020.

\bibitem{gatys2016image}
Leon~A Gatys, Alexander~S Ecker, and Matthias Bethge.
\newblock Image style transfer using convolutional neural networks.
\newblock In {\em Proceedings of the IEEE conference on computer vision and
  pattern recognition}, pages 2414--2423, 2016.

\bibitem{han2017deepsketch2face}
Xiaoguang Han, Chang Gao, and Yizhou Yu.
\newblock Deepsketch2face: a deep learning based sketching system for 3d face
  and caricature modeling.
\newblock {\em ACM Transactions on graphics (TOG)}, 36(4):1--12, 2017.

\bibitem{han2018caricatureshop}
Xiaoguang Han, Kangcheng Hou, Dong Du, Yuda Qiu, Shuguang Cui, Kun Zhou, and
  Yizhou Yu.
\newblock Caricatureshop: Personalized and photorealistic caricature sketching.
\newblock {\em IEEE transactions on visualization and computer graphics},
  26(7):2349--2361, 2018.

\bibitem{he2016deep}
Kaiming He, Xiangyu Zhang, Shaoqing Ren, and Jian Sun.
\newblock Deep residual learning for image recognition.
\newblock In {\em Proceedings of the IEEE conference on computer vision and
  pattern recognition}, pages 770--778, 2016.

\bibitem{isola2017image}
Phillip Isola, Jun-Yan Zhu, Tinghui Zhou, and Alexei~A Efros.
\newblock Image-to-image translation with conditional adversarial networks.
\newblock In {\em Proceedings of the IEEE conference on computer vision and
  pattern recognition}, pages 1125--1134, 2017.

\bibitem{johnson2016perceptual}
Justin Johnson, Alexandre Alahi, and Li Fei-Fei.
\newblock Perceptual losses for real-time style transfer and super-resolution.
\newblock In {\em European conference on computer vision}, pages 694--711.
  Springer, 2016.

\bibitem{FairFace}
Kimmo Karkkainen and Jungseock Joo.
\newblock Fairface: Face attribute dataset for balanced race, gender, and age
  for bias measurement and mitigation.
\newblock In {\em Proceedings of the IEEE/CVF Winter Conference on Applications
  of Computer Vision}, pages 1548--1558, 2021.

\bibitem{ffhq}
Tero Karras, Samuli Laine, and Timo Aila.
\newblock A style-based generator architecture for generative adversarial
  networks.
\newblock In {\em Proceedings of the IEEE/CVF conference on computer vision and
  pattern recognition}, pages 4401--4410, 2019.

\bibitem{kim2021k}
Taewoo Kim, Chaeyeon Chung, Sunghyun Park, Gyojung Gu, Keonmin Nam, Wonzo Choe,
  Jaesung Lee, and Jaegul Choo.
\newblock K-hairstyle: A large-scale korean hairstyle dataset for virtual hair
  editing and hairstyle classification.
\newblock In {\em 2021 IEEE International Conference on Image Processing
  (ICIP)}, pages 1299--1303. IEEE, 2021.

\bibitem{lewiner2011interactive}
Thomas Lewiner, Thales Vieira, Dimas Mart{\'\i}nez, Adelailson Peixoto,
  Vin{\'\i}cius Mello, and Luiz Velho.
\newblock Interactive 3d caricature from harmonic exaggeration.
\newblock {\em Computers \& Graphics}, 35(3):586--595, 2011.

\bibitem{li2021deep}
Song Li, Songzhi Su, Juncong Lin, Guorong Cai, and Li Sun.
\newblock Deep 3d caricature face generation with identity and structure
  consistency.
\newblock {\em Neurocomputing}, 454:178--188, 2021.

\bibitem{lin2021meingame}
Jiangke Lin, Yi Yuan, and Zhengxia Zou.
\newblock Meingame: Create a game character face from a single portrait.
\newblock In {\em Proceedings of the AAAI Conference on Artificial
  Intelligence}, volume~35, pages 311--319, 2021.

\bibitem{CelebA}
Ziwei Liu, Ping Luo, Xiaogang Wang, and Xiaoou Tang.
\newblock Deep learning face attributes in the wild.
\newblock In {\em Proceedings of International Conference on Computer Vision
  (ICCV)}, December 2015.

\bibitem{metahuman}
Unreal engine metahuman.
\newblock \url{https://www.unrealengine.com/en-US/metahuman}.
\newblock Accessed: 2022-11-10.

\bibitem{novelai}
Novelai.
\newblock \url{https://novelai.net/}.
\newblock Accessed: 2022-11-10.

\bibitem{park2020contrastive}
Taesung Park, Alexei~A Efros, Richard Zhang, and Jun-Yan Zhu.
\newblock Contrastive learning for unpaired image-to-image translation.
\newblock In {\em European Conference on Computer Vision}, pages 319--345.
  Springer, 2020.

\bibitem{paszke2017automatic}
Adam Paszke, Sam Gross, Soumith Chintala, Gregory Chanan, Edward Yang, Zachary
  DeVito, Zeming Lin, Alban Desmaison, Luca Antiga, and Adam Lerer.
\newblock Automatic differentiation in pytorch.
\newblock 2017.

\bibitem{peng2017parametric}
Weilong Peng, Zhiyong Feng, Chao Xu, and Yong Su.
\newblock Parametric t-spline face morphable model for detailed fitting in
  shape subspace.
\newblock In {\em Proceedings of the IEEE Conference on Computer Vision and
  Pattern Recognition}, pages 6139--6147, 2017.

\bibitem{richardson2021encoding}
Elad Richardson, Yuval Alaluf, Or Patashnik, Yotam Nitzan, Yaniv Azar, Stav
  Shapiro, and Daniel Cohen-Or.
\newblock Encoding in style: a stylegan encoder for image-to-image translation.
\newblock In {\em IEEE/CVF Conference on Computer Vision and Pattern
  Recognition (CVPR)}, June 2021.

\bibitem{AgileAvatar}
Shen Sang, Tiancheng Zhi, Guoxian Song, Minghao Liu, Chunpong Lai, Jing Liu,
  Xiang Wen, James Davis, and Linjie Luo.
\newblock Agileavatar: Stylized 3d avatar creation via cascaded domain
  bridging.
\newblock {\em ACM SIGGRAPH Asia 2022 Conference Proceedings}, 2022.

\bibitem{F2P}
Tianyang Shi, Yi Yuan, Changjie Fan, Zhengxia Zou, Zhenwei Shi, and Yong Liu.
\newblock Face-to-parameter translation for game character auto-creation.
\newblock In {\em Proceedings of the IEEE/CVF International Conference on
  Computer Vision}, pages 161--170, 2019.

\bibitem{shi2020fast}
Tianyang Shi, Zhengxia Zuo, Yi Yuan, and Changjie Fan.
\newblock Fast and robust face-to-parameter translation for game character
  auto-creation.
\newblock In {\em Proceedings of the AAAI Conference on Artificial
  Intelligence}, volume~34, pages 1733--1740, 2020.

\bibitem{song2021agilegan}
Guoxian Song, Linjie Luo, Jing Liu, Wan-Chun Ma, Chunpong Lai, Chuanxia Zheng,
  and Tat-Jen Cham.
\newblock Agilegan: stylizing portraits by inversion-consistent transfer
  learning.
\newblock {\em ACM Transactions on Graphics (TOG)}, 40(4):1--13, 2021.

\bibitem{svanera2016figaro}
Michele Svanera, Umar~Riaz Muhammad, Riccardo Leonardi, and Sergio Benini.
\newblock Figaro, hair detection and segmentation in the wild.
\newblock In {\em 2016 IEEE International Conference on Image Processing
  (ICIP)}, pages 933--937. IEEE, 2016.

\bibitem{maad1}
Philipp Terh{\"{o}}rst, Daniel F{\"{a}}hrmann, Jan~Niklas Kolf, Naser Damer,
  Florian Kirchbuchner, and Arjan Kuijper.
\newblock Maad-face: {A} massively annotated attribute dataset for face images.
\newblock {\em {IEEE} Trans. Inf. Forensics Secur.}, 16:3942--3957, 2021.

\bibitem{maad2}
Philipp Terh{\"{o}}rst, Marco Huber, Jan~Niklas Kolf, Ines Zelch, Naser Damer,
  Florian Kirchbuchner, and Arjan Kuijper.
\newblock Reliable age and gender estimation from face images: Stating the
  confidence of model predictions.
\newblock In {\em 10th {IEEE} International Conference on Biometrics Theory,
  Applications and Systems, {BTAS} 2019, Tampa, FL, USA, September 23-26,
  2019}, pages 1--8. {IEEE}, 2019.

\bibitem{tov2021designing}
Omer Tov, Yuval Alaluf, Yotam Nitzan, Or Patashnik, and Daniel Cohen-Or.
\newblock Designing an encoder for stylegan image manipulation.
\newblock {\em ACM Transactions on Graphics (TOG)}, 40(4):1--14, 2021.

\bibitem{vieira2013three}
Roberto C~Cavalcante Vieira, Creto~A Vidal, and Joaquim~Bento Cavalcante-Neto.
\newblock Three-dimensional face caricaturing by anthropometric distortions.
\newblock In {\em 2013 XXVI Conference on Graphics, Patterns and Images}, pages
  163--170. IEEE, 2013.

\bibitem{cifar-n}
Jiaheng Wei, Zhaowei Zhu, Hao Cheng, Tongliang Liu, Gang Niu, and Yang Liu.
\newblock Learning with noisy labels revisited: A study using real-world human
  annotations.
\newblock In {\em International Conference on Learning Representations}, 2022.

\bibitem{wu2018alive}
Qianyi Wu, Juyong Zhang, Yu-Kun Lai, Jianmin Zheng, and Jianfei Cai.
\newblock Alive caricature from 2d to 3d.
\newblock In {\em Proceedings of the IEEE Conference on Computer Vision and
  Pattern Recognition}, pages 7336--7345, 2018.

\bibitem{xu2020deep}
Sicheng Xu, Jiaolong Yang, Dong Chen, Fang Wen, Yu Deng, Yunde Jia, and Xin
  Tong.
\newblock Deep 3d portrait from a single image.
\newblock In {\em Proceedings of the IEEE/CVF Conference on Computer Vision and
  Pattern Recognition}, pages 7710--7720, 2020.

\bibitem{Yang_2020_CVPR}
Haotian Yang, Hao Zhu, Yanru Wang, Mingkai Huang, Qiu Shen, Ruigang Yang, and
  Xun Cao.
\newblock Facescape: A large-scale high quality 3d face dataset and detailed
  riggable 3d face prediction.
\newblock In {\em Proceedings of the IEEE/CVF Conference on Computer Vision and
  Pattern Recognition (CVPR)}, June 2020.

\bibitem{ye20213d}
Zipeng Ye, Mengfei Xia, Yanan Sun, Ran Yi, Minjing Yu, Juyong Zhang, Yu-Kun
  Lai, and Yong-Jin Liu.
\newblock 3d-carigan: an end-to-end solution to 3d caricature generation from
  normal face photos.
\newblock {\em IEEE Transactions on Visualization and Computer Graphics}, 2021.

\bibitem{yin2017learning}
Weidong Yin, Yanwei Fu, Yiqing Ma, Yu-Gang Jiang, Tao Xiang, and Xiangyang Xue.
\newblock Learning to generate and edit hairstyles.
\newblock In {\em Proceedings of the 25th ACM international conference on
  Multimedia}, pages 1627--1635, 2017.

\bibitem{zhu2017unpaired}
Jun-Yan Zhu, Taesung Park, Phillip Isola, and Alexei~A Efros.
\newblock Unpaired image-to-image translation using cycle-consistent
  adversarial networks.
\newblock In {\em Proceedings of the IEEE international conference on computer
  vision}, pages 2223--2232, 2017.

\bibitem{zhu2021mind}
Peihao Zhu, Rameen Abdal, John Femiani, and Peter Wonka.
\newblock Mind the gap: Domain gap control for single shot domain adaptation
  for generative adversarial networks, 2021.

\bibitem{zollhofer2018state}
Michael Zollh{\"o}fer, Justus Thies, Pablo Garrido, Derek Bradley, Thabo
  Beeler, Patrick P{\'e}rez, Marc Stamminger, Matthias Nie{\ss}ner, and
  Christian Theobalt.
\newblock State of the art on monocular 3d face reconstruction, tracking, and
  applications.
\newblock In {\em Computer Graphics Forum}, volume~37, pages 523--550. Wiley
  Online Library, 2018.

\end{thebibliography}
